\documentclass{article}
\usepackage[T1]{fontenc}
\usepackage[utf8]{inputenc}
\PassOptionsToPackage{table}{xcolor}
\usepackage{main}
\usepackage{microtype}
\usepackage{graphicx}
\usepackage{float}
\usepackage{newtxtext}  
\usepackage{amsmath}
\usepackage{amssymb}
\usepackage{newtxmath}  
\usepackage{booktabs}
\definecolor{mydarkblue}{rgb}{0,0.08,0.45}
\usepackage[colorlinks=true,linkcolor=mydarkblue,citecolor=mydarkblue,filecolor=mydarkblue,urlcolor=mydarkblue]{hyperref}
\usepackage{fancyhdr}
\fancypagestyle{titlepage}{%
  \fancyhf{}%
  \lhead{\raisebox{-0.8cm}[0.65cm][0.3cm]{\includegraphics[height=0.95cm]{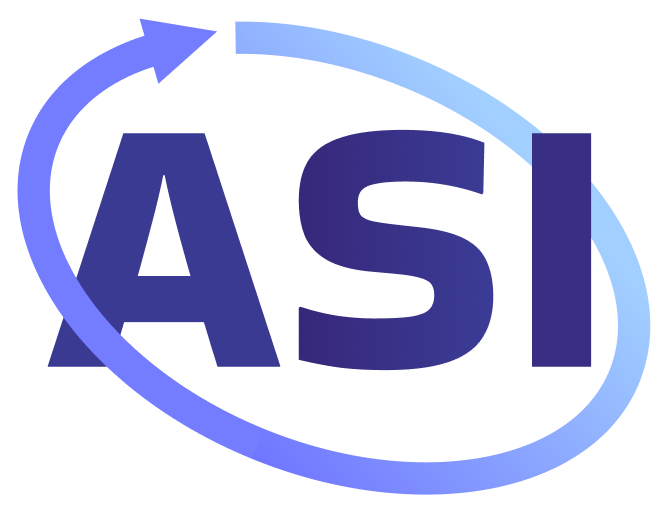}}}%
  \rhead{\raisebox{-0.8cm}[0.5cm][0.2cm]{\includegraphics[height=0.7cm]{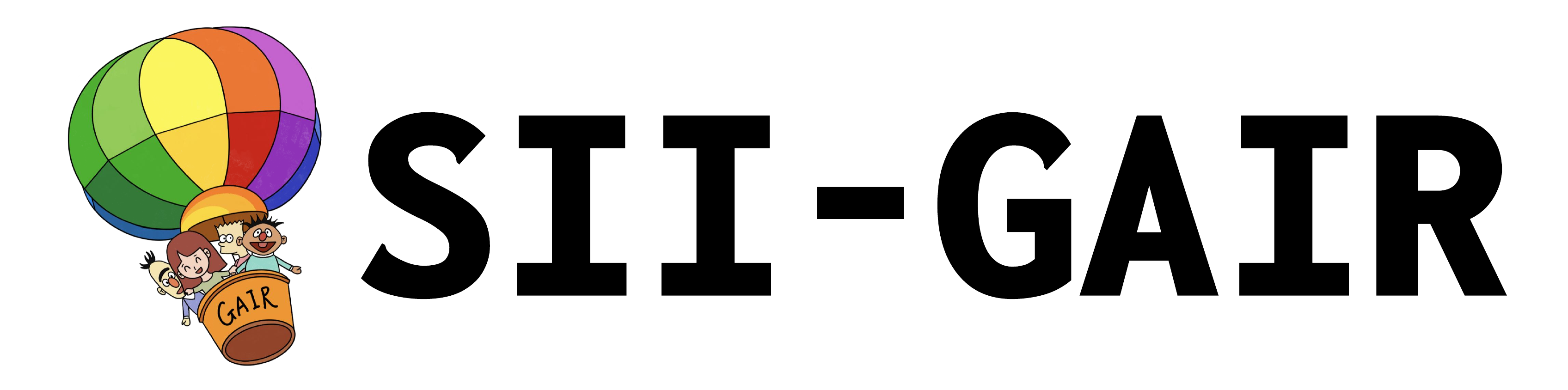}}}%
  \cfoot{\thepage}%
}

\fancypagestyle{normal}{%
  \fancyhf{}%
  \lhead{ERPBench}%
  \rhead{\raisebox{-0.08cm}{\includegraphics[height=0.55cm]{assets/GAIR_logo_sii.pdf}}}%
  \cfoot{\thepage}%
}

\DeclareCaptionFont{black}{\color{black}}

\usepackage{etoolbox}
\newcounter{bibcount}
\makeatletter
\patchcmd{\@lbibitem}{\item[}{\item[\hfil\stepcounter{bibcount}{[\thebibcount]}}{}{}
\renewcommand\NAT@bibsetup%
  [1]{\setlength{\leftmargin}{\bibhang}\setlength{\itemindent}{-\parindent}%
      \setlength{\itemsep}{\bibsep}\setlength{\parsep}{\z@}}
\makeatother

\makeatletter
\renewcommand{\@toptitlebar}{%
  {\color{black}\hrule height 1\p@}%
  \vskip 0.25in
  \vskip -\parskip%
}
\renewcommand{\@maketitle}{%
  \vbox{%
    \hsize\textwidth
    \linewidth\hsize
    \vskip 0.1in
    \@toptitlebar
    \vskip 0.1in
    {\centering\LARGE\bfseries \@title\par}%
    \vskip 0.15in
    {\raggedright\@author\par}%
    \vskip 0.3in \@minus 0.1in
  }
}
\renewenvironment{abstract}%
  {\centerline{\large\bfseries Abstract}%
   \begin{list}{}%
     {\setlength{\rightmargin}{0.6cm}%
      \setlength{\leftmargin}{0.6cm}}%
   \item[]\ignorespaces}%
  {\unskip\end{list}\vspace{1.5ex}}
\def\section{\@startsection {section}{1}{\z@}{-2.0ex plus
    -0.5ex minus -.2ex}{1.5ex plus 0.3ex minus .2ex}{\large\bfseries\raggedright}}
\def\subsection{\@startsection{subsection}{2}{\z@}{-1.8ex plus
    -0.5ex minus -.2ex}{0.8ex plus .2ex}{\normalsize\bfseries\raggedright}}
\def\subsubsection{\@startsection{subsubsection}{3}{\z@}{-1.5ex plus
   -0.5ex minus -.2ex}{0.5ex plus .2ex}{\normalsize\bfseries\raggedright}}
\def\paragraph{\@startsection{paragraph}{4}{\z@}{1.5ex plus
   0.5ex minus .2ex}{-1em}{\normalsize\bfseries}}
\makeatother

\begin{document}

\title{ERPBench: Evaluating LLM Agents for Enterprise Decision-Making Across Competitive Market Ecologies}
\author{}

\maketitle
\thispagestyle{titlepage}

\begin{center}
\textbf{Xinran Zhang}\textsuperscript{1,2,4,*} \quad
\textbf{Pengrui Lu}\textsuperscript{1,3,4,*} \quad
\textbf{Lyumanshan Ye}\textsuperscript{3,4} \quad
\textbf{Pengfei Liu}\textsuperscript{1,3,4,\textdagger}\\[3pt]
{\small
\textsuperscript{1}Shanghai Innovation Institute \quad
\textsuperscript{2}Beijing Institute of Technology\\
\textsuperscript{3}Shanghai Jiao Tong University \quad
\textsuperscript{4}GAIR Lab\\[2pt]
\textsuperscript{*}Equal contribution. \quad
\textsuperscript{\textdagger}Corresponding author.}
\end{center}

\begin{abstract}
Large language model (LLM) agents are increasingly proposed for enterprise workflows, yet existing evaluations rarely test whether business-decision conclusions transfer across competitive market ecologies. We introduce ERPBench, an execution-instrumented benchmark for enterprise decision agents in a six-round Enterprise Resource Planning (ERP) simulation with coupled pricing, production, procurement, inventory, finance, and shared-market competition. ERPBench evaluates the same 100 fixed problems in two matched competitive market ecologies: \emph{Solo}, where each evaluated LLM agent competes against fixed rule-based opponents, and \emph{Arena}, where six evaluated LLM agents compete in a shared market. Across six model families, this yields 1,200 model-level trajectories spanning 7,200 decision rounds. Under the observed service configuration, the leading model differs between ecologies: DeepSeek leads in \emph{Solo} (252.29M mean valuation; mean rank 1.67), whereas Gemini leads in \emph{Arena} (263.95M; 1.76). The two ecologies identify the same task-level winner on only 21 of 100 problems, and Gemini's bottom-rank rate falls from 22\% to 0\% in \emph{Arena}. ERPBench supports paired evaluation of whether enterprise-agent rankings transfer across competitive market ecologies, supplemented by aggregate execution-intervention analysis. Code and benchmark resources are available in our \href{https://github.com/GAIR-NLP/erp-bench}{GitHub repository}.
\end{abstract}

\section{Introduction}

LLM-agent evaluation has moved from static question answering toward stateful digital environments, including tool use, code repair, web navigation, desktop control, and workplace workflows~\cite{liu2023agentbench,jimenez2024swebench,zhou2024webarena,xie2024osworld,yao2024taubench,xu2025agentcompany}. Enterprise decision agents require a complementary form of reliability: they repeatedly allocate scarce resources while pricing, production, procurement, inventory, and finance decisions compound over time. In competitive markets, these decisions are also strategically coupled, because one agent's price or inventory policy changes the demand landscape faced by the others. We call the composition and behavior of opponents that share a market with an evaluated LLM agent the \emph{competitive market ecology} (or \emph{ecology} for short). Thus, the relevant benchmark question is not only whether an agent can produce a high-scoring business plan, but whether conclusions about the agent remain stable when the same business problem is embedded in a different ecology.

Formally, ERPBench asks: \emph{does the selected model differ across matched competitive market ecologies for the same enterprise task?} To answer it, we need three measurement ingredients. First, a \textbf{primary outcome metric}: terminal company valuation computed by a dividend discount model (DDM) at the end of six decision rounds. Second, \textbf{complementary reliability indicators}: mean within-problem rank, bottom-rank rate (the fraction of problems on which a model ranks last), and normalized regret (the gap to the best model scaled by the within-problem valuation range). Third, an \textbf{execution audit layer}: logged decision records and execution-audit information, where available, linking parsed decisions, feasibility checks, clamping, fallback, and runtime recovery so that a terminal score can be qualified by how it was produced. Three structural challenges make this measurement non-trivial: (1)~\emph{multi-round compounding}---decisions span six rounds of thirty simulated days each, so early pricing or production errors propagate forward; (2)~\emph{joint constraint satisfaction}---pricing, production, procurement, inventory, and finance decisions are mutually constrained by capacity, cash, and demand, so a locally plausible action may be globally infeasible; and (3)~\emph{strategic coupling under competition}---in a shared market, one agent's price or inventory policy changes every other agent's demand, so identical decisions can yield different outcomes against different opponents.

\begin{figure}[H]
\centering
\includegraphics[width=0.96\textwidth]{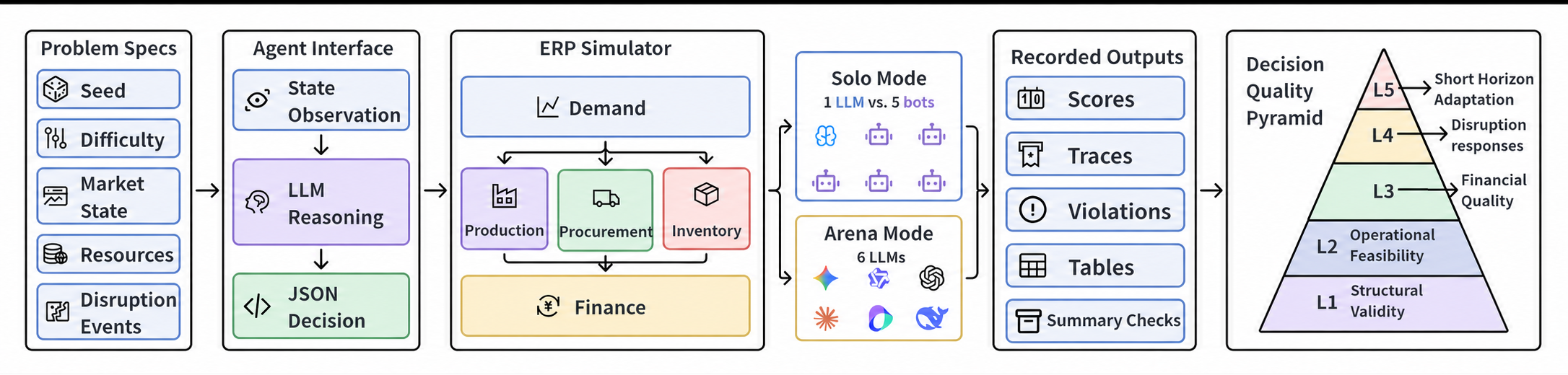}
\caption{ERPBench overview. The same 100 ERP problems are evaluated under matched \emph{Solo} and \emph{Arena} competitive market ecologies. The protocol matches problem configurations and model-family labels while changing the opponent composition; six decision rounds yield terminal outcomes and execution-audit records where available.}
\label{fig:overview}
\end{figure}

A conventional benchmark design would fix a single competitive market ecology and report one terminal leaderboard. This design is attractive because it is simple and reproducible, but it is incomplete for enterprise competition: opponent behavior is not background noise. It changes prices, demand allocation, inventory pressure, market share, and downstream cash-flow outcomes. As a result, a model can be strong against stable rule-based competitors yet lose its advantage when other evaluated LLM agents shape the same market; conversely, a model with weak isolated performance may become more reliable in a shared market. Enterprise-agent benchmarks should therefore treat competitive market ecology as a controlled measurement condition rather than incidental setup.

A second challenge is that enterprise decisions must be executable, not merely plausible. A model may return a syntactically valid action that violates production capacity, requests unavailable investment, over-orders under cash constraints, or requires fallback parsing. Terminal valuation alone does not reveal whether the score was produced through direct execution or through repeated repair. For this reason, an enterprise benchmark should record parsed decisions, feasibility checks, clamping, fallback, simulation updates, and terminal outcomes.

We introduce ERPBench, an execution-instrumented benchmark for measuring enterprise decision agents under controlled competitive market ecologies. ERPBench uses an independently constructed Enterprise Resource Planning (ERP) simulation covering production, supply chain, inventory, finance, market competition, and carbon costs. Each of the same 100 ERP problems is evaluated under two matched ecologies: \emph{Solo}, where each evaluated LLM agent competes against fixed rule-based opponents, and \emph{Arena}, where six evaluated LLM agents compete in a shared market. The paired design matches problem instances and model-family labels across ecologies, yielding 1,200 model-level trajectories spanning 7,200 decision rounds across six model families.

ERPBench makes three contributions. First, it provides a controlled ERP-agent task substrate in which pricing, production, procurement, inventory, and finance decisions are jointly constrained over a six-round horizon. Second, its central measurement contribution is a paired ecology protocol that evaluates the same model-family--problem pairs under fixed-opponent \emph{Solo} and shared-market \emph{Arena} conditions. Third, it supplements outcome comparisons with logged execution-intervention evidence: DeepSeek leads \emph{Solo} while Gemini leads \emph{Arena}, the two ecologies identify the same task-level winner on only 21 of 100 problems, and tail-risk profiles differ.

The resulting benchmark treats both competitive market ecology and execution-audit information as part of the measurement specification. This goes beyond a terminal score: for ERP agents, the scoring path includes whether a structured action could be parsed, whether it respected capacity, procurement, inventory, and finance constraints, and whether the simulator required repair, clamping, fallback, or runtime recovery. ERPBench therefore supports not only outcome comparison, but also process-level characterization of how outcomes were produced.

\section{Related Work}

\textbf{Stateful executable agent benchmarks.}
Recent agent benchmarks have moved evaluation beyond static question answering toward interactive, stateful, and executable tasks. AgentBench, SWE-bench, WebArena, OSWorld, and $\tau$-bench evaluate agents in environments where actions change state and outcomes can be checked through code, web interfaces, desktop applications, or tool-mediated user interactions~\cite{liu2023agentbench,jimenez2024swebench,zhou2024webarena,xie2024osworld,yao2024taubench}. This line of work establishes a central baseline for ERPBench: agent evaluation should be reproducible, stateful, and outcome-verifiable. However, these benchmarks usually hold the surrounding competitive market ecology fixed. ERPBench asks a complementary question: whether conclusions about the same enterprise decision task remain stable when the competitive market ecology changes.

\textbf{Enterprise and professional agent benchmarks.}
A second line of work evaluates agents in professional software and workplace settings. TheAgentCompany and WorkArena++ study agents completing realistic workplace tasks, while CRMArena and CRMArena-Pro move this direction into business-facing CRM environments with realistic enterprise data, roles, and interactions~\cite{xu2025agentcompany,boisvert2024workarenaplusplus,huang2025crmarena,huang2025crmarena_pro}. These benchmarks are important because they test whether agents can operate inside professional workflows rather than isolated toy tasks. ERPBench differs in the object of measurement: it evaluates coupled enterprise resource planning decisions rather than CRM or office-task completion. In ERPBench, pricing, production, procurement, inventory, capacity, finance, and final valuation are jointly constrained by the simulator, so a plausible plan is not sufficient unless it can be executed.

\textbf{Economic and competitive agents.}
Economic-agent benchmarks study LLM behavior in markets, financial decisions, retail operations, and competitive environments. CompeteAI, EconAgent, InvestorBench, RetailBench, and Market-Bench provide evidence that LLM agents can be evaluated as economic decision-makers rather than only language or tool users~\cite{chen2024competeai,li2024econagent,li2025investorbench,zhang2026retailbench,zheng2026marketbench}. ERPBench builds on this direction but changes the measurement target. Its central comparison is not only which model performs best in a business domain, but whether the same model--problem pair yields the same conclusion under the fixed-opponent \emph{Solo} ecology and the shared-market \emph{Arena} ecology populated by evaluated LLM agents. This paired design makes competitive market ecology an explicit experimental factor.

\textbf{Benchmark validity, trajectories, and protocol sensitivity.}
Recent work also argues that benchmark conclusions depend on evaluation design, interaction protocol, and trajectory evidence. $\tau^2$-Bench studies dynamic multi-party trajectories~\cite{barres2025tau2bench}, benchmark-sensitivity analyses show that model rankings can shift under protocol changes~\cite{alzahrani2024benchmarktargets}, and benchmark-design studies such as BetterBench emphasize validity, reproducibility, and lifecycle quality in evaluation artifacts~\cite{reuel2024betterbench}. ERPBench operationalizes these concerns for enterprise agents by retaining round-level decision records and execution-audit information, where available, and reporting paired outcome changes across controlled competitive market ecologies. The audit records show not only what an agent proposed, but also whether the action could be parsed, repaired, clamped, or executed under ERP constraints. In this sense, ERPBench treats both competitive market ecology and execution-audit information as part of the measurement specification.

\begin{figure}[H]
    \centering
    \includegraphics[width=0.88\textwidth]{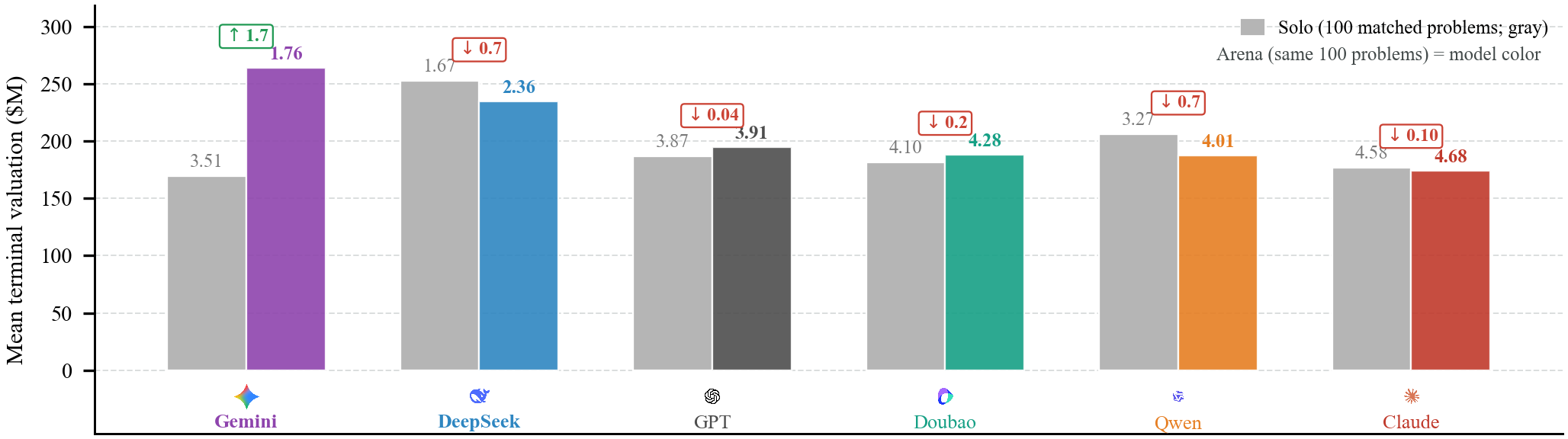}
    \caption{Aggregate outcomes under matched competitive market ecologies. Gray bars: \emph{Solo} mean valuation across 100 matched problems; colored bars: \emph{Arena} mean valuation on the same problems. Numbers above bars report mean rank (lower is better). Arrows show rank-shift magnitude and direction (green improves; red worsens).}
    \label{fig:solo-arena-panel}
\end{figure}

\section{ERPBench}

\subsection{Operational Simulation and Executable Interface}

ERPBench is designed around three measurement requirements. First, the task must be operationally coupled: pricing, production, procurement, inventory, capacity, cash, and valuation should affect one another rather than appear as isolated subtasks. Second, decisions must be executable: an agent should not receive credit for a plausible business plan unless it can be parsed and run under ERP constraints. Third, the benchmark must support ecological comparison: the same model-family--problem pair should be measurable under different competitive market ecologies while holding the underlying task fixed.

To instantiate these requirements, ERPBench builds on ERPsim-style business games~\cite{leger2007erpsim,leger2011erpsim} and simulates a cereal manufacturing firm over six rounds of 30 simulated days each. Agents decide SKU prices, production quantities, procurement, marketing allocation, and financial actions for 12 SKUs across three distribution channels. The environment includes batch production with capacity and lot-size constraints, supplier lead times, warehouse inventory, cash-flow and credit dynamics, carbon costs, and a multinomial competitive interaction (MCI) demand model~\cite{cooper1988mci}. These components make terminal performance depend on both strategic choices and operational feasibility: a pricing plan that ignores inventory, for example, can increase nominal margin while losing demand or creating stockouts.

Each evaluated model family uses the same single-agent scaffold and the same executable interface. Before each decision, the LLM agent receives the current ERP state and may query tools for finance, inventory, market price, demand, raw materials, and carbon information. It then returns a structured JSON action containing SKU prices, production quantities, purchase orders, marketing allocation, and finance actions. The simulator parses the action, checks feasibility, repairs or clamps infeasible fields when necessary, executes the round, and records the resulting ERP state. Where execution audit logs are available, they contain both the proposed decision and the interventions required to make that decision executable.

This interface is intentionally stricter than a natural-language planning task. ERP decisions are not merely recommendations; they must survive capacity, cash, inventory, and availability constraints. The action space is also deliberately coupled. Prices influence demand through the competitive market model, but demand can only be converted into value if the agent has produced and stocked the right SKUs in the right channels. Procurement choices affect future production rather than only the current round, and finance actions can preserve cash viability but may not be available in every state. These dependencies allow early mistakes to propagate through inventory, capacity, and cash-flow constraints before appearing in the final valuation.

\subsection{Paired Ecologies and Measurement Protocol}

ERPBench evaluates each model family on the same 100 ERP problem instances under two matched competitive market ecologies. In \emph{Solo}, the focal LLM agent competes against five deterministic rule-based opponents with conservative, aggressive, balanced, growth-oriented, and risk-averse policies. In \emph{Arena}, six evaluated LLM agents compete in a shared market. The matched comparison preserves the problem identifier, seed, difficulty stratum, scenario category, agent scaffold, decision horizon, outcome metrics, and model-family label while changing the opponent composition. \emph{Solo} ranks are computed by comparing the six model families' separate fixed-opponent trajectories on the same problem; rule-based opponents do not enter the reported model ranking. We use \emph{LLM agent} for a deployed decision-making instance that receives ERP state and tool outputs and returns structured multi-round decisions; \emph{model family} denotes the underlying evaluated LLM condition, and \emph{model} is shorthand only when reporting rankings. Each model--problem--ecology trajectory spans six rounds.

The primary outcome is sixth-round terminal company valuation, which aggregates profitability, financial viability, and expected future cash flows. We also report within-problem rank, task-level winner agreement, normalized regret, bottom-rank rate, difficulty-stratified effects, and logged execution-intervention categories. For each model $m$ and problem $p$, the paired ecology effect is
\begin{equation}
\Delta_{p,m}=V^{\mathrm{Arena}}_{p,m}-V^{\mathrm{Solo}}_{p,m}.
\end{equation}
We report model-level mean effects with 95\% percentile-bootstrap confidence intervals over problems. Paired valuation tests use two-sided Wilcoxon signed-rank tests with Benjamini--Hochberg correction across the six model-wise tests. Logged execution-intervention categories are non-exclusive process diagnostics, not causal decompositions of final valuation. For the descriptive reliability--performance panel, the \emph{L2 operational reliability score} for model family $m$ in \emph{Arena} is $1-r_m/\max_{m'} r_{m'}$, where $r_m$ is its logged operational-intervention rate per model-level trajectory. This within-panel normalization is an audit summary, not a primary outcome metric or causal measure. The Decision Quality Pyramid in Figure~\ref{fig:overview} is an organizing taxonomy, not a composite score: L1 structural validity summarizes parser-related interventions; L2 operational feasibility is summarized by the operational reliability score; L3 financial quality is rank-derived terminal performance; L4 disruption response restricts that performance to hard and random-hard tasks; and L5 short-horizon adaptation is the average improvement in within-problem rank from Round 1 to Round 6. L3--L5 are descriptive proxies.

The 100 problems cover baseline, dual-pressure, and harder operational scenarios. We use these strata to check whether an ecology effect is robust across problem families rather than driven by a single scenario type. This matters because an ERP benchmark can otherwise reward a model that specializes in one high-variance slice while being unreliable in routine or constraint-heavy settings. ERPBench therefore reports both aggregate performance and structured heterogeneity across task families.

\section{Experiments}

\subsection{Setup}

We evaluate six model families---Claude Opus 4.6, DeepSeek-V4-Flash, Gemini 3.1 Pro Preview, GPT-5.5, Doubao Seed 2.0 Pro, and Qwen 3.7 Max---under the paired ecology protocol described above. All models were evaluated with the same agent scaffold and six-round protocol.

The analysis uses the same 100 problem identifiers in both \emph{Solo} and \emph{Arena}. Each ecology contains 600 model-level terminal outcome records and 3,600 ordered model-level decision-round records. Thus, the benchmark contains 1,200 model-level trajectories and 7,200 model-level decision rounds. Operationally, these correspond to 600 independent \emph{Solo} simulator executions and 100 shared six-model \emph{Arena} executions, or 700 simulator executions in total. Terminal valuations are read from the sixth-round decision record. Coverage, duplicate, and missing-record checks confirm 100/100 problem coverage, no duplicate model--problem records, and no missing expected records.

This design makes the comparisons below paired rather than cross-sectional. When we report that a model gains or loses valuation in \emph{Arena}, the quantity is computed on the same problem set with the same model-family label and problem seed. It is therefore a paired benchmark contrast under the observed service configuration, not a comparison between separately sampled leaderboards.

\subsection{Ecology Reorders Aggregate Performance}
\label{sec:aggregate-results}

\begin{figure}[!htbp]
\centering
\begin{minipage}[t]{0.55\textwidth}
\vspace{0pt}
\small
\raggedright
Table~\ref{tab:aggregate-mode-effects} and Figure~\ref{fig:solo-arena-panel} show that the aggregate leaderboard changes across ecologies. DeepSeek leads \emph{Solo} with 252.29M mean valuation and mean rank 1.67, whereas Gemini leads \emph{Arena} with 263.95M and mean rank 1.76. Because the comparison is paired by problem, the reversal is not a difference between unrelated samples: Gemini gains +94.54M on the same problems, while DeepSeek and Qwen decrease by 17.58M and 18.74M, respectively.

The benchmark implication is that a single-ecology leaderboard gives an incomplete conclusion: \emph{Solo} selects DeepSeek, while \emph{Arena} selects Gemini. ERPBench exposes this discrepancy by treating competitive market ecology as a measurement variable rather than assuming one model is universally superior.

\par\medskip
\centering
\captionsetup{type=table}
\captionof{table}{Matched-ecology aggregate performance. Val. is mean terminal valuation (M); lower rank is better; $\Delta V$ is \emph{Arena}--\emph{Solo}. Confidence intervals appear in Appendix~\ref{app:statistics}.}
\label{tab:aggregate-mode-effects}
\scriptsize
\renewcommand{\arraystretch}{1.08}
\setlength{\tabcolsep}{1.2pt}
\begin{tabular*}{\linewidth}{@{\extracolsep{\fill}}lrrrrrr@{}}
\toprule
\textbf{Model} & \multicolumn{2}{c}{\textbf{Solo}} & \multicolumn{2}{c}{\textbf{Arena}} & \textbf{$\Delta V$} & \textbf{$q$} \\
\cmidrule(lr){2-3}\cmidrule(lr){4-5}\cmidrule(lr){6-7}
& Val. & Rank & Val. & Rank & M & \\
\midrule
DeepSeek & \textbf{252.29} & \textbf{1.67} & 234.71 & 2.36 & $-17.58$ & $.0116$ \\
Gemini & 169.41 & 3.51 & \textbf{263.95} & \textbf{1.76} & \textbf{$+94.54$} & \textbf{$1.28{\times}10^{-15}$} \\
GPT-5.5 & 186.43 & 3.87 & 194.82 & 3.91 & $+8.39$ & $.126$ \\
Qwen & 205.87 & 3.27 & 187.13 & 4.01 & $-18.74$ & $.0061$ \\
Doubao & 181.27 & 4.10 & 188.17 & 4.28 & $+6.90$ & $.557$ \\
Claude & 176.78 & 4.58 & 173.95 & 4.68 & $-2.83$ & $.866$ \\
\bottomrule
\end{tabular*}
\end{minipage}
\hfill
\begin{minipage}[t]{0.42\textwidth}
\vspace{0pt}
\centering
\includegraphics[width=\linewidth]{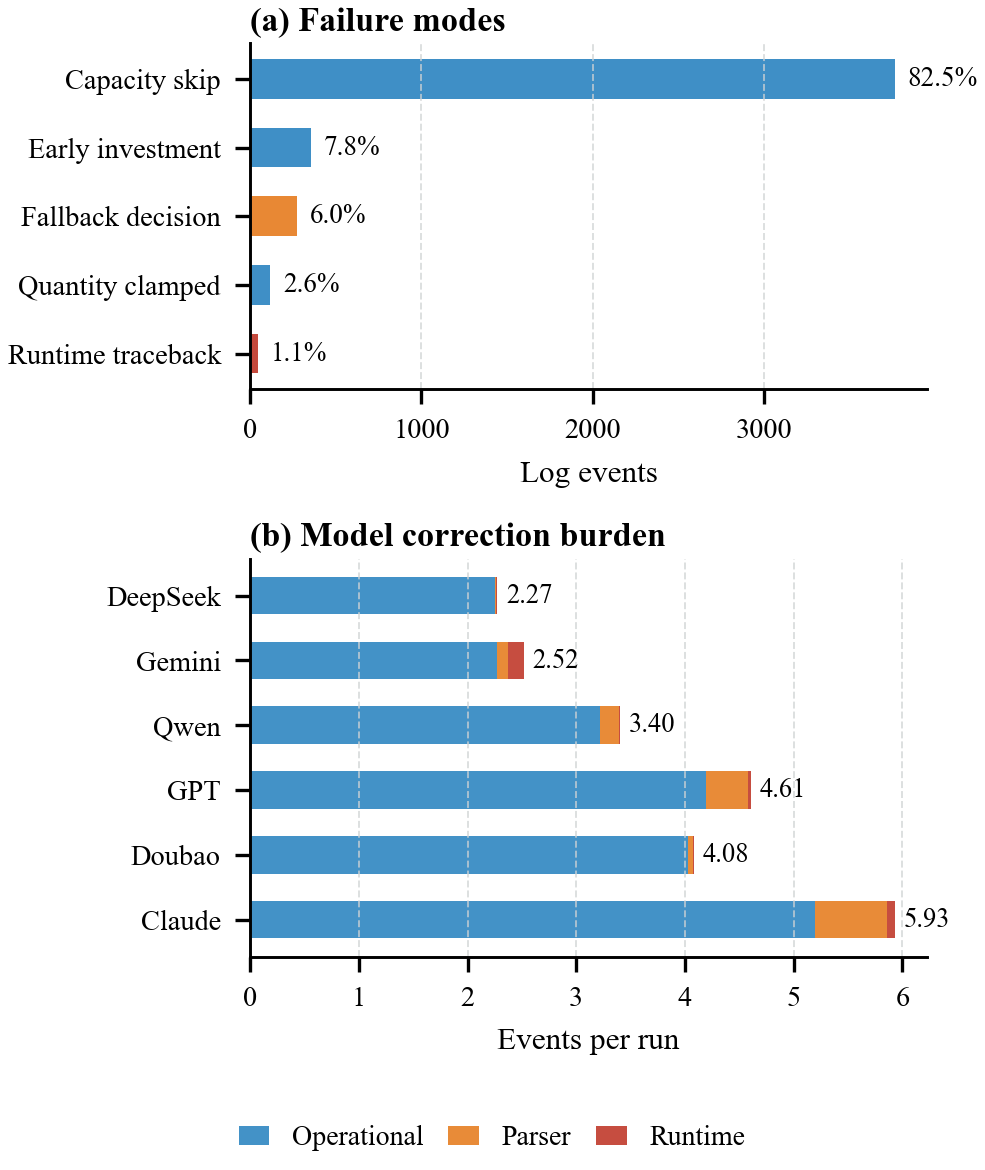}
\captionsetup{justification=raggedright,singlelinecheck=false}
\captionof{figure}{Execution-intervention summary. Most logged events are production-capacity skips; fallback decisions and tracebacks indicate output-pipeline or runtime recovery.}
\label{fig:trace-intervention-profile}
\end{minipage}
\end{figure}

\subsection{Terminal Scores Need Trace Qualification}

A terminal valuation is necessary for ranking ERP agents, but it is not sufficient for diagnosing enterprise-agent reliability. The same final score can be reached through a clean executable path or through repeated parsing recovery, feasibility repair, clamping, or fallback. These two cases are both valid simulator outcomes, but they imply different reliability profiles for deployment.

ERPBench therefore uses execution-audit records behind terminal valuations where logging is available. These records link model outputs, parsing results, feasibility checks, repair or clamping events, fallback paths, executed actions, and resulting ERP states. The audit layer lets the analysis separate three questions that would otherwise be conflated: which model obtains the highest value, whether that conclusion transfers across competitive market ecologies, and whether aggregate performance coincides with substantial execution-intervention burden.

\subsection{Execution Audit Exposes Execution Fragility}

Logged execution-audit records show where the executable decision pipeline required intervention. Among the five displayed intervention types in Figure~\ref{fig:trace-intervention-profile}, production-capacity skips account for the largest share (3,761 events, 82.5\%), followed by unavailable investment choices (355, 7.8\%), fallback decisions (275, 6.0\%), production-quantity clamping (118, 2.6\%), and tracebacks (48, 1.1\%). Panel (b) pools \emph{Solo} and \emph{Arena} trajectories, whereas Figure~\ref{fig:reliability-performance-coupling} uses \emph{Arena}-only rates. The displayed shares exclude five omitted run-level failures.

These logged categories should be interpreted as process diagnostics rather than causal decompositions of terminal valuation. A capacity skip often means that a requested production item could not be scheduled under machine capacity, lot-size, or material constraints; it may reflect an aggressive production plan in a constrained plant rather than a malformed decision. By contrast, fallback decisions and tracebacks more directly indicate output-pipeline or runtime fragility. Execution audit therefore qualifies the aggregate outcome measurement by summarizing how often the executable pipeline required intervention.

\subsection{Task-Level Rankings Disagree}

The aggregate reversal is not only a mean-score artifact. At the individual-task level, the two ecologies identify the same task-level winner on only 21 of 100 problems, and the mean within-problem Spearman rank correlation is 0.182 with bootstrap CI [0.090, 0.273]. The largest transition is from DeepSeek to Gemini: 35 problems won by DeepSeek in \emph{Solo} are won by Gemini in \emph{Arena}.

\begin{figure}[H]
\centering
\includegraphics[width=0.82\textwidth]{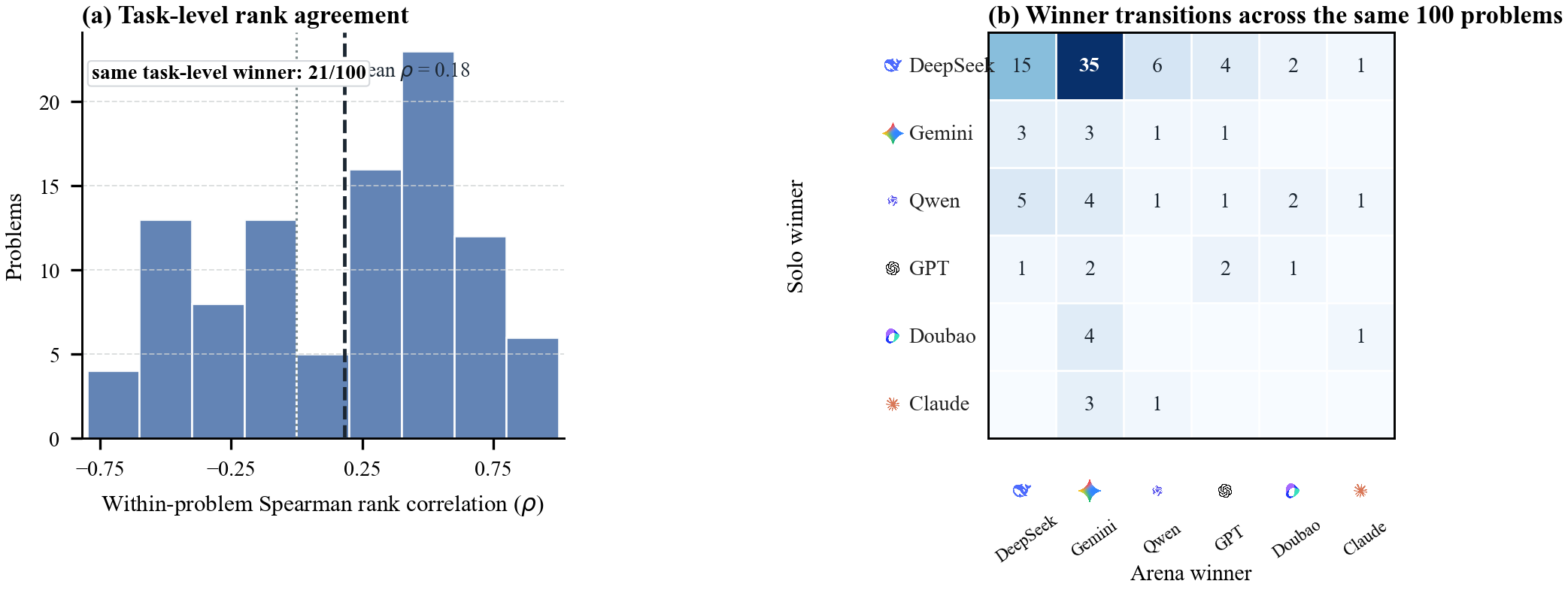}
\caption{Task-level disagreement between \emph{Solo} and \emph{Arena} across the same 100 matched ERP problems. The two ecologies identify the same task-level winner on only 21 problems and often produce different rankings.}
\label{fig:task-level-disagreement}
\end{figure}

Figure~\ref{fig:task-level-disagreement} shows that ecology changes both the identity of the best model and the rank ordering within many problems. This matters for benchmark interpretation. If the aggregate reversal were driven by a few extreme outliers, it would be weaker evidence of ecology-sensitive measurement. Broad winner disagreement instead shows that the competitive market ecology changes many local judgments about which model is best for the same ERP task.

\subsection{Reliability Qualifies Terminal Scores}

\begin{figure}[H]
\centering
\includegraphics[width=0.92\textwidth]{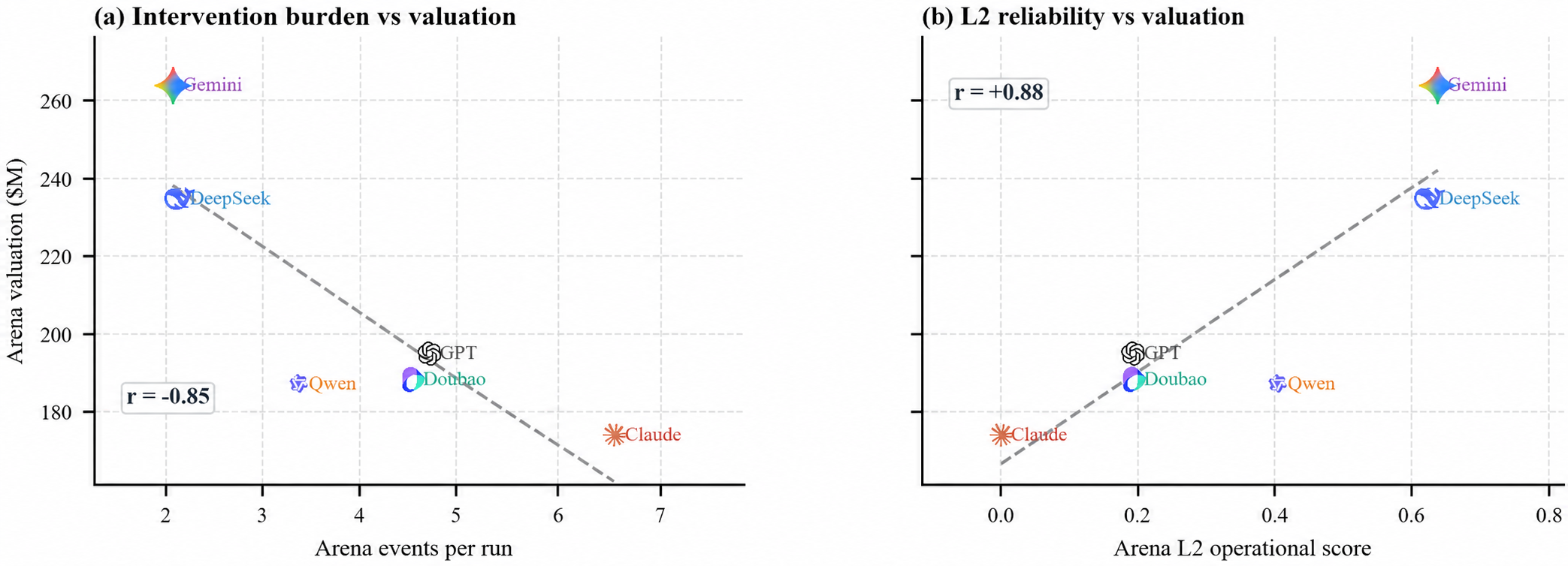}
\caption{Descriptive reliability--performance patterns under the \emph{Arena} ecology across six model families. (a)~Logged execution-intervention burden versus mean terminal valuation; (b)~L2 operational reliability score versus mean terminal valuation. The L2 score is a within-panel normalization of logged operational-intervention rates; these six-point associations are descriptive, not inferential.}
\label{fig:reliability-performance-coupling}
\end{figure}

Ecology also changes lower-tail reliability. Gemini's bottom-rank rate falls from 22\% in \emph{Solo} to 0\% in \emph{Arena}, and its P90 normalized regret falls from 0.73 to 0.27. DeepSeek shows the opposite tail pattern, with P90 regret rising from 0.16 to 0.39. Figure~\ref{fig:reliability-performance-coupling} provides a descriptive view of execution-audit intervention burden and terminal valuation: models with fewer logged interventions tend to achieve higher Arena valuations, yet the six-model relationship admits substantial variation and is not inferential.

This result gives the paired ecology protocol a deployment-oriented interpretation. \emph{Solo} measures whether a model can independently produce high-quality plans against stable opponents. \emph{Arena} measures whether the same model remains competitive when peer evaluated LLM agents change prices, demand allocation, and market outcomes. A model with high average valuation but frequent lower-tail collapse may be less reliable than its mean score suggests; conversely, a model that becomes stable in a shared market may exhibit strengths missed by fixed opponents.

\subsection{Difficulty Conditions the Ecology Effect}

\begin{figure}[H]
\centering
\includegraphics[width=0.74\textwidth]{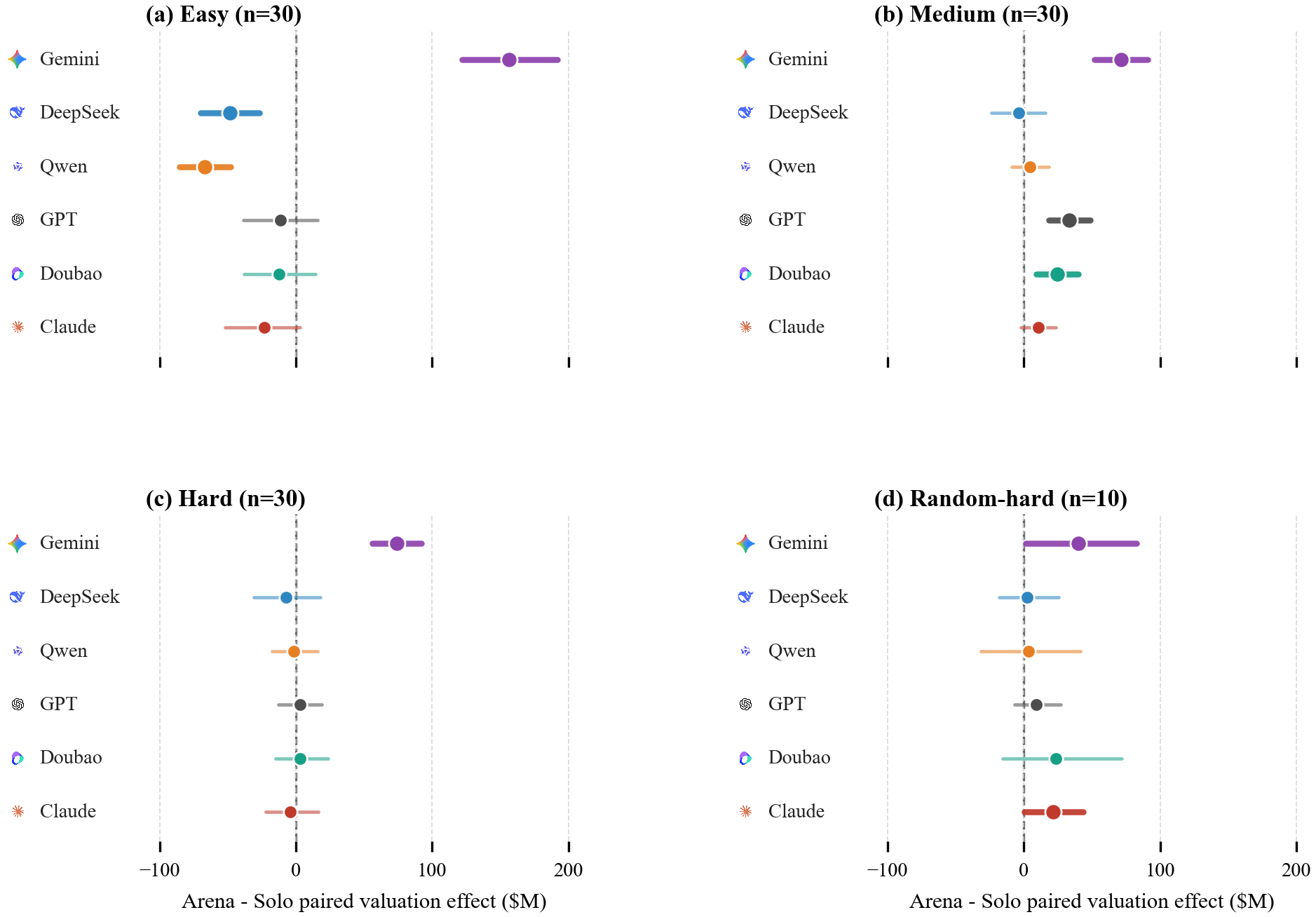}
\caption{Difficulty-stratified paired effects. Ecology effects are heterogeneous across task strata, showing that the \emph{Arena} shift is not a uniform offset applied to every problem type. Points show paired mean effects; horizontal lines show 95\% bootstrap confidence intervals; bold intervals exclude zero.}
\label{fig:difficulty-mode-effects}
\end{figure}

Difficulty-stratified analysis tests whether the ecology effect is a single global shift or a task-conditioned change. Figure~\ref{fig:difficulty-mode-effects} shows that the paired \emph{Arena}--\emph{Solo} effects vary across difficulty strata. Gemini consistently benefits across the displayed strata, whereas the other model families show smaller, mixed-direction shifts; uncertainty also varies with stratum size and operational difficulty.

This heterogeneity is important for benchmark validity. If the result were caused only by a few extreme easy or hard cases, the benchmark would provide weaker evidence about enterprise-agent reliability. Instead, the stratified view shows that the direction of the main ecology effect is visible across the problem design, while the scale of the effect depends on the operational pressure imposed by the task. ERPBench should therefore be read as a structured problem-family benchmark rather than a single undifferentiated score pool.

\section{Discussion and Limitations}

ERPBench shows that single-ecology leaderboards can be incomplete for enterprise agents. \emph{Solo} measures performance against stable rule-based competitors; \emph{Arena} measures behavior when other evaluated LLM agents also shape prices, demand allocation, inventory, and market outcomes. Neither ecology is universally superior or uniquely correct. The paired comparison instead tests whether conclusions about a model transfer across explicitly specified competitive market ecologies.

The empirical pattern suggests two different forms of enterprise-agent competence. DeepSeek appears strongest as an independent optimizer: it leads \emph{Solo}, has very low lower-tail exposure in that setting, and often produces high-quality decisions against fixed opponents. Gemini appears strongest as a competitive responder: it has higher tail risk in \emph{Solo}, but becomes the clear aggregate leader in \emph{Arena} and avoids bottom-rank outcomes. This distinction would be invisible in a benchmark that reports only one competitive market ecology.

Outcome panels and execution-audit summaries play complementary roles. Terminal valuation and rank establish the benchmark comparison, while aggregate audit summaries characterize parsing recovery, feasibility repair, clamping, and runtime recovery. Logged interventions should not be overread as causal mechanisms or interchangeable model-error units; they constrain interpretation and support process-level analysis. In particular, production-capacity skips can reflect a binding environment constraint rather than a malformed decision, whereas fallbacks and tracebacks are stronger evidence of output-pipeline fragility.

The ecology effect also varies by problem: models with near-neutral aggregate changes can improve on some tasks and worsen on others. It is therefore a problem-dependent interaction, not a constant shift applied to each model.
ERPBench remains a controlled ERP-inspired simulation, with a fixed 100-problem slice, six-round horizon, and one six-model \emph{Arena} composition; results may differ under additional ecologies, fixed-opponent sets, longer horizons, or information regimes. Service-side routing and model endpoints may change over time, so model-family labels do not guarantee immutable or route-identical vendor snapshots across ecologies. The paired results should therefore be read as benchmark contrasts under the observed service configuration rather than as route-invariant causal effects. Finally, logged interventions are audit evidence rather than a mechanism model. Future work should test candidate mechanisms involving price following, capacity, inventory, and market share.

\section{Conclusion}

ERPBench evaluates enterprise decision agents under paired competitive market ecologies with logged intervention evidence. Across 100 matched ERP problems, six model families, and 1,200 model-level trajectories, the leading model changes from DeepSeek in \emph{Solo} to Gemini in \emph{Arena}; task-level winners, difficulty effects, and tail-risk profiles also change. These results suggest that enterprise-agent evaluation should report not only terminal scores, but also whether model-selection conclusions transfer across competitive market ecologies and how executable decisions are produced.
\noindent\textit{Additional details, including the problem-set coverage audit, the full confidence-interval table, and the trace taxonomy, are provided in Appendices~\ref{app:coverage}--\ref{app:provenance}.}

\section{Acknowledgments}
We thank Professor Senchun Chai for valuable suggestions on the ERP system and its system-level design. We thank Professor Qiuju Yin of the School of Management, Beijing Institute of Technology, for supporting system testing, sharing the SAP competition rules, and providing strategic feedback that helped shape this project. We also thank the Beijing Institute of Technology students who participated in the SAP ERPSIM competition for their early testing and practical feedback on ERP-market competition.

{\small
\bibliographystyle{acl_natbib}
\bibliography{related}
}

\clearpage
\appendix
\small

\section{Problem Set and Coverage Audit}
\label{app:coverage}

The 100-problem slice is assembled as a matched benchmark panel rather than as independent ad hoc runs. Each problem specifies a seed, difficulty stratum, scenario category, initial operational state, and market parameters. The \emph{Solo100} and \emph{Arena100} panels use the same problem identifiers so that every model--problem pair has a direct ecology comparison. Each ecology contains 600 terminal outcome records, corresponding to 100 problems times six evaluated model families. Internal consistency checks found 100/100 problem coverage, zero duplicate model--problem records, and zero missing expected records.

The main paper reports terminal valuations from the sixth-round record. During analysis, ordered per-round records were used to associate model identity, problem identifier, round number, parsed decisions, feasibility events, and resulting simulation states. Execution-intervention summaries were computed from logged events where available.

The consistency checks cover problem--model coverage, duplicate pairs, missing expected records, and source-identifier associations where identifiers are present. Coverage requires each problem to appear for all six models, while duplicate and missing-record checks enforce exactly 600 terminal outcome records per ecology. These checks support the paired design without changing the terminal valuation or ranking definitions.

\section{Metrics and Statistical Protocol}
\label{app:statistics}

Valuation is reported in millions. Mean rank is computed within each problem and ecology among the six evaluated model families, with lower rank indicating better terminal valuation. Normalized regret is computed within a problem as the gap to the best model under the same ecology, scaled by the observed within-problem valuation range. Bottom-rank rate is the fraction of problems on which a model ranks last among the evaluated families.

\begin{table}[!htbp]
\centering
\small
\renewcommand{\arraystretch}{1.12}
\setlength{\tabcolsep}{8pt}
\caption{Paired ecology effects with 95\% confidence intervals and Benjamini--Hochberg-adjusted $q$-values. Values match Table~\ref{tab:aggregate-mode-effects}; intervals are percentile-bootstrap confidence intervals over problems.}
\label{tab:appendix-effects}
\begin{tabular}{lrrr}
\toprule
\textbf{Model} & $\Delta V$ (M) & \textbf{95\% CI (M)} & $q$ \\
\midrule
DeepSeek & $-17.58$ & [$-29.79$, $-5.36$] & $.0116$ \\
Gemini & $+94.54$ & [$+78.74$, $+110.40$] & $1.28{\times}10^{-15}$ \\
GPT-5.5 & $+8.39$ & [$-2.84$, $+19.54$] & $.126$ \\
Qwen & $-18.74$ & [$-29.89$, $-7.56$] & $.0061$ \\
Doubao & $+6.90$ & [$-4.90$, $+19.04$] & $.557$ \\
Claude & $-2.83$ & [$-14.60$, $+8.42$] & $.866$ \\
\bottomrule
\end{tabular}
\end{table}

For paired effects, the unit of resampling is the problem, not the individual row. This preserves the matched design because each resampled problem contributes the corresponding \emph{Solo} and \emph{Arena} outcomes for the same model. Statistical tests use two-sided Wilcoxon signed-rank tests on paired valuation differences and apply Benjamini--Hochberg correction across the six model-wise tests. We treat these tests as descriptive evidence for ecology sensitivity rather than as a claim about all possible ERP markets. The same principle applies to pairwise win rates and winner-disagreement summaries: they are benchmark statistics for this controlled 100-problem panel, not estimates of universal model superiority. Their role is to show whether the paired protocol produces stable or unstable conclusions under a defined measurement condition.

\section{Trace Taxonomy and Audit Scope}
\label{app:provenance}

Trace interventions are divided into six logged categories. \emph{Production-capacity skip} denotes an attempted production item that could not be scheduled under available machine capacity, lot-size, or material constraints. \emph{Investment unavailable} denotes a requested financial action that was not available in the current state. \emph{Fallback decision} denotes recovery from missing or malformed model output into a valid default action. \emph{Production quantity clamped} denotes a numeric production field adjusted to an allowed range. \emph{Traceback} denotes runtime recovery captured by the execution log, and \emph{experiment failed} denotes a run-level execution failure. Figure~\ref{fig:trace-intervention-profile} displays the five most frequent categories; the five \emph{experiment failed} events are omitted and excluded from the displayed-share denominator.

These categories are intentionally not summed into a single ``error rate.'' They occur at different layers of the pipeline and may overlap. Capacity skips are often normal consequences of the ERP environment, whereas fallback and traceback events are closer to interface robustness failures. We therefore use them only as aggregate process diagnostics that qualify, rather than replace or causally explain, the terminal outcome analysis.

\normalsize
\end{document}